\documentclass[lettersize,journal]{IEEEtran}

\usepackage{amssymb,amsmath}
\usepackage{cite}
\usepackage{subcaption}
\usepackage{graphicx}
\usepackage{psfrag}
\usepackage{url}
\usepackage[latin1]{inputenc}
\usepackage[absolute,overlay]{textpos}
\usepackage{gensymb}
\usepackage{cases}
\usepackage{bm}
\usepackage{graphicx}
\usepackage[font=footnotesize]{caption}
\usepackage[font=footnotesize]{subcaption}
\usepackage{float}
\usepackage[linesnumbered,ruled,lined]{algorithm2e}

\usepackage{tabularx}
\usepackage{array}

\allowdisplaybreaks
\usepackage{tikz}
\usepackage{booktabs,multirow}
\usetikzlibrary{patterns,arrows}
\definecolor{color1}{RGB}{237, 191, 193}
\definecolor{color2}{RGB}{229, 153, 157}
\definecolor{color3}{RGB}{225, 123, 116}
\definecolor{color4}{RGB}{10,10,200}
\definecolor{color5}{RGB}{203,52,38}

\usepackage{pgf}

\usepackage[nocomma]{optidef}

\usepackage{algorithmic}

\usepackage{multicol}
\usepackage{amsthm}

\usepackage{booktabs}
\usepackage{color,soul}

\begin{document}

\title{{\it MIRA}: A \textbf{M}ethod of Federated Mult\textbf{I}-Task  Learning for La\textbf{R}ge L\textbf{A}nguage Models

}

\author{
	\IEEEauthorblockN{Ahmed Elbakary$^{1}$, Chaouki Ben Issaid$^1$, Tamer ElBatt$^{3,4}$, Karim Seddik$^2$, Mehdi Bennis$^1$ \\
	}
	\IEEEauthorblockA{$^1$Centre for Wireless Communications (CWC), University of Oulu, Finland\\
	$^2$Department of Electronics and Communications Engineering, American University in Cairo, Egypt\\
        $^3$Department of Computer Science and Engineering, American University in Cairo, Egypt\\ 
        $^4$Department of Electronics and Communications Engineering, Cairo University, Egypt 
    }
    }
\maketitle
\begin{abstract}
In this paper, we introduce a method for fine-tuning Large Language Models (LLMs), inspired by Multi-Task learning in a federated manner. Our approach leverages the structure of each client's model and enables a learning scheme that considers other clients' tasks and data distribution. To mitigate the extensive computational and communication overhead often associated with LLMs, we utilize a parameter-efficient fine-tuning method, specifically Low-Rank Adaptation (LoRA), reducing the number of trainable parameters. Experimental results, with different datasets and models, demonstrate the proposed method's effectiveness compared to existing frameworks for federated fine-tuning of LLMs in terms of average and local performances. The proposed scheme outperforms existing baselines by achieving lower local loss for each client while maintaining comparable global performance.

\end{abstract}


\section{Introduction}
Pre-trained Large Language Models (LLMs) are proven to be few-shot learners~\cite{radford2019language}, which means that they can perform a diverse set of tasks without retraining on those specific tasks. This property is attributed mainly to two factors: the wide range of their training data and the scale of the architecture. While in-context learning~\cite{min2022rethinking} might be enough to adapt the LLM to the task at hand, this might not be possible without further model fine-tuning. The problem arises when adapting these models to a domain-specific area where the data comes from a different distribution than the training data itself. This requires further fine-tuning of the models so they can perform at a reasonable level. Gathering such domain-specific data might be impractical because of the associated cost of data collection in addition to privacy concerns that might arise. 

Federated learning (FL)~\cite{mcmahan2017communication} is one technique that can be leveraged to solve such a problem without revealing local data. The idea is to enable different clients to train a model jointly to solve a specific problem by leveraging local data for local training and sharing only the model's parameters, not the local data. Traditional FL algorithms aim to train a global model that can be deployed across all participating clients, regardless of the underlying distribution of their data. Multiple frameworks already exist to leverage FL for fine-tuning LLMs. For example, Zhang \textit{et al.} \cite{zhang2023fedpetuning}~proposed the usage of different Parameter-efficient fine-tuning (PEFT) methods like prompt tuning to make the fine-tuning process of LLMs communication efficient. The idea of prompt tuning is to add a set of virtual tokens, often called a soft prompt, to the original prompt, i.e., the hard prompt. Then, it tries to optimize the soft prompt to generate the desired output. In \cite{xu2023federated}, the authors made use of the fact that most FL settings operate with edge devices designed by nature for inference not training. This characteristic leads to solving the problem using a zeroth-order method without the need for backpropagation (BP). Another zeroth-order method that shares only two numbers, the scalar gradient and the seed used to generate the perturbation vector necessary to calculate the gradient, was proposed in \cite{malladi2023fine}. Although this approach is communication-efficient, it introduces a computation overhead since each client needs to perform several local steps to get the latest model. 

The main problem with the existing federated fine-tuning approaches for LLMs is the fact that they all learn a single global model based on model averaging, which might not be the optimal solution for highly heterogeneous settings. However, this might not be the optimal solution when the model is expected to perform well across different tasks where the data distribution differs significantly. Federated Multi-Task learning (FMTL)~\cite{smith2017federated} provides a promising framework that can be leveraged to solve the issue of heterogeneous clients. Several approaches have been proposed to adapt MTL to model FL problems. For example, in \cite{li2021ditto}, the authors aligned the local learned model and the global one using regularization improving, both, model personalization and fairness among clients. The existence of a hidden relationship between clients' models was assumed in \cite{9975151}, and the Laplacian matrix was leveraged to capture this relationship.
\begin{figure}[t]
    \centering
    \includegraphics[scale=0.3]{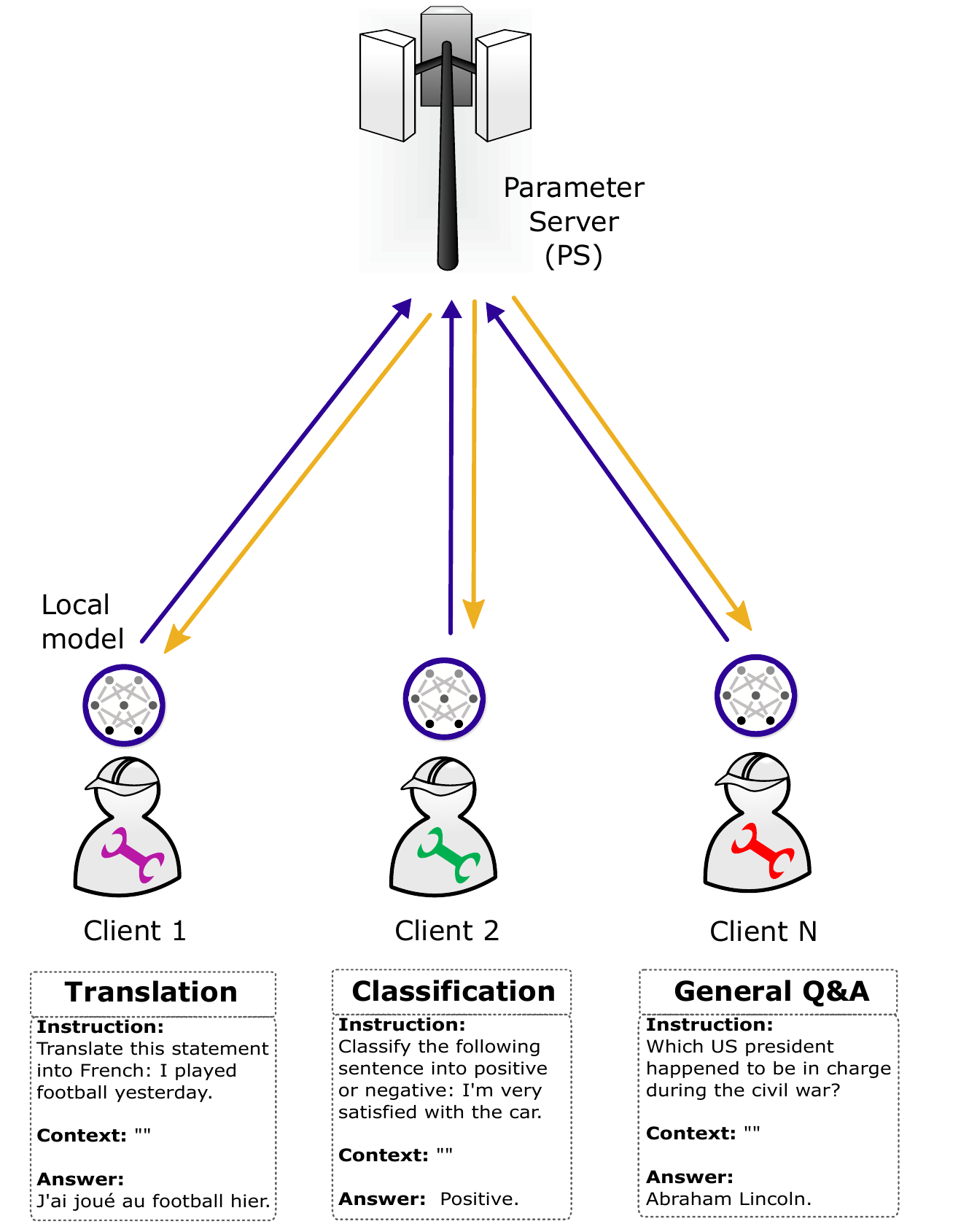} 
        \caption{A schematic illustration of the FMTL framework for fine-tuning LLMs. Each client is equipped with the same pre-trained model and a different task.}
        \label{fig:tasks}
\end{figure}
The second issue is concerned with the computation and communication costs associated with LLMs. An LLM is, at its core, a very deep neural network, with millions of weights. Most state-of-the-art techniques for training neural networks utilize gradient-based methods like stochastic gradient descent (SGD). An SGD step requires computing the gradient of the loss function with respect to (w.r.t) the model's parameters using BP. However, BP suffers from one major bottleneck, which is the memory footprint needed to compute the gradient. Along with storing models' parameters, gradient-based methods require the activation of individual neurons in the network to be stored, enabling the optimizer to compute a backward pass, which is needed to compute the gradient and the SGD update. Even worse, some modern optimizers need to store additional information, like the exponential moving average of the gradients. This comes with almost a minimal cost in the case of shallow or deep networks but not very deep networks like LLMs. In fact, in the case of a billion-sized network, like an LLM, storing both the model's parameters and the activation values would imply that tens of gigabytes are needed for such a model. Since those kinds of models are usually trained using graphics processing units (GPUs), the cost might be quite high for even a very limited training time and might take days or weeks unless a very powerful GPU cluster is available. 
One approach to solve this issue is by substituting the original weight matrix with a lower-dimensional matrix that can be trained while freezing the original weight matrix. PEFT methods seek a middle ground by adding a small set of parameters to the pre-trained model and only training the newly added parameters. Low-Rank Adaptation (LoRA)~\cite{hu2021LoRA} is one technique that enables such decomposition, where the original weight matrix is decomposed into two sub-matrices. The two matrices are then initialized with Gaussian and zero initialization, respectively. During training, the original weight matrix is frozen, while LoRA's weights are the only trainable parameters to be fine-tuned. This technique for fine-tuning ensures that the stored activation values and the optimizer's state are minimal in comparison to that of the original model. We leverage both the FMTL paradigm and PEFT and apply them to federating the process of finetuning LLMs. The contributions of this work are multi-fold
\begin{itemize}
    \item We propose FMTL as a new paradigm for fine-tuning LLMs in a federated manner to address the issue of data heterogeneity among clients and enable each client to learn a custom model more suitable for its data distribution.
    \item We leverage a PEFT mechanism, specifically LoRA, to significantly reduce the computational and communication overhead associated with LLMs, ensuring efficient and scalable fine-tuning.
    \item Extensive evaluation is carried out using two different LLMs: Data-juicer~\cite{chen2023data} and GPT2-large~\cite{radford2019language}, and two datasets: Natural Instructions (NI) ~\cite{wang2022super} and Dolly15-k~\cite{DatabricksBlog2023DollyV2}, to show the effectiveness of the proposed scheme.
\end{itemize}

\section{System Model and Problem Formulation}
We assume a network of $K$ clients, with the set of all clients defined as $\mathcal{N} = \{1, \dots, K\}$ and $\mathcal{N}_k = \mathcal{N} \setminus \{k\}$ being the set of all neighboring clients for client $k$. Clients communicate only with the parameter server (PS), but not with each other directly. Each client trains a separate model $\bm{w}_k \in \mathbb{R}^{d}$, according to its own local data $(\bm{X}_k, \bm{y}_k)$, where $\bm{X}_k$ represents the feature matrix, the instruction to the LLM in our case, and $\bm{y}_k$ represents the label vectors, the expected output from the LLM. Along with that, clients have their own tasks ${t_k}$ and objective function $f_k({\bm{w}_k}): \mathbb{R}^d \xrightarrow{} \mathbb{R}$, $\forall k \in \mathcal{N}$. The setting is depicted in Figure~\ref{fig:tasks}. We model the interactions among clients as a connected graph $\mathcal{G}$, with an adjacency matrix  $\bm{M}$ that quantifies the similarity between tasks of different clients. The entries of the matrix $\bm{M}$ represent how strong the relationship is between two clients, i.e., the degree of similarity between the two tasks. 

Let $\bm{W} = [\bm{w}_{1}^{T}, \dots, \bm{w}_{K}^{T}] \in \mathbb{R}^{dK}$ be a concatenation of all clients' models and let $\bm{D} = \operatorname{diag}\left[\delta_1, \delta_2, \ldots, \delta_K\right] \in \mathbb{R}^{K \times K}$ be a diagonal matrix that represents the summation of all neighboring clients connections where $\delta_k = \sum_{\ell \in \mathcal{N}_k} a_{k\ell}$. The Laplacian matrix of the graph is defined as $\bm{L} = \bm{D} - \bm{M}$, and its extended version as $\bm{\mathcal{L}} = \bm{L} \otimes \bm{I}_{d}$, where $\otimes$ denotes the Kronecker product of two matrices and $\bm{I}_d$ is the identity matrix. The FMTL problem aims to minimize the overall objective function, which comprises the global loss and a regularization term enforcing task similarity as follows
\begin{align}
    \min_{\bm{W}} J(\bm{W}) = F(\bm{W}) + \lambda R(\bm{W}),
\end{align}
where $\lambda$ is a regularization hyperparameter which controls how much weight a client gives to other clients' models. In other words, if $\lambda = 0$, the problem is converted into a traditional FL problem with full local training and no collaboration between clients is taken into account. On the other hand if $\lambda > 0$, we encourage similar tasks/clients to align their models close to each other. The global loss $F(\bm{W})$ is defined as the sum of local losses across all clients
\begin{align}
    F(\bm{W}) = \sum_{k=1}^{K} f_k(\bm{w}_k).
\end{align}
\begin{figure}[t]
        \includegraphics[width=0.9\linewidth]{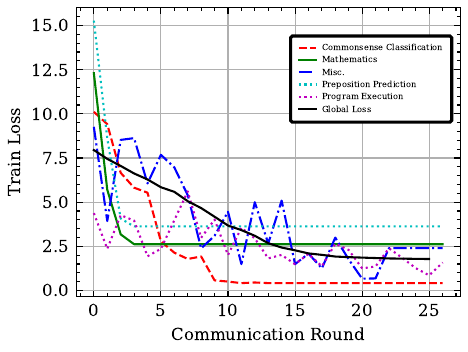} 
        \caption{Global model versus per-task model performance.}
        \label{fig:onecolumn}
\end{figure}
The Laplacian regularization term $\mathcal{R}(\bm{W})$ captures the similarity between clients' task models and is defined as 
\begin{align}\label{reg.term}
    \mathcal{R}(\bm{W})=\bm{W}^T \mathcal{L} \bm{W}=\frac{1}{2} \sum_{k=1}^N \sum_{\ell \in \mathcal{N}_k} a_{k \ell}\left\|\bm{w}_k-\bm{w}_{\ell}\right\|^2.
\end{align}
In this formulation, $f_k(\bm{w}_k)$ represents the local loss function at client $k$, computed by evaluating a sample $\bm{x}_k^{(i)}$ using the model $\bm{w}_k$. The norm $\left\| \cdot \right\|$ denotes the Euclidean norm, and $a_{k\ell}$ quantifies the similarity between tasks of clients $k$ and $\ell$. In other words, with higher values of $a_{k, \ell}$, it indicates stronger relationships. The objective function $J(\bm{W})$ consists of two components: the global loss, which is optimized locally by each client, and the regularization term, which is handled by the server.

To minimize $J(\bm{W})$, we perform both local and global updates iteratively. At each client $k$, we perform $R$ local optimization steps to update $\bm{w}_k$. After completing the local updates, the server minimizes the regularization term $\mathcal{R}(\bm{W})$ by adjusting the clients' models based on their similarities. Specifically, the server updates each client's model as follows
\begin{align}\label{reg.update}
     \bm{w}_k^{(t+1)} = \bm{w}_{k, R}^{(t)} - \eta \lambda \sum_{\ell \in \mathcal{N}_k} a_{k\ell} (\bm{w}_{k, R}^{(t)} - \bm{w}_{\ell, R}^{(t)}), \forall k \in \mathcal{N},
\end{align} 
where $\bm{w}_{k, R}^{(t)}$ is the locally updated model of client $k$ after $R$ steps at iteration $t$, and $\eta$ is the server's learning rate.

To motivate the FMTL setting, we conduct a preliminary experiment with the DataJuicer LLM~\cite{chen2023data} on the NI dataset \cite{wang2022super}. We measure the performance of the global learned model on different tasks by reporting the average loss per task using one of the latest proposed methods, FedKSeed~\cite{qin2023federated} against learning a fully local model. As we can see from Figure \ref{fig:onecolumn}, the performance of the model varies across tasks. Most importantly, the global model's performance is worse than the local ones in certain tasks with a high variance in the case where the size of the local dataset is small. This can be attributed to the fact that learning a single global model, as in the traditional FL approach, is sub-optimal in the case of different tasks or data distributions across clients. On the other hand, learning a fully local model is near optimal in case sufficient compute and data are available, which is not the case for most edge devices. To address the limitation of the traditional FL approach, we hypothesize that learning a separate model per client can solve this problem, enabling clients to handle their local tasks better while still taking into account the structure and the similarity of other clients' tasks.

\newcommand{\RegularizedUpdate}[1]{%
  \textbf{Function} RegularizedUpdate($\mathcal{H}$)
  \FOR{each $<s_{\ell, r}, g_{\ell, r}>$ \in \mathcal{H}$}
  \STATE Generate perturbation vector $z{_\ell, r}$ based on s.
  \STATE Compute the regularized update as in \eqref{update-multiple-t}.
  \ENDFOR
}

\begin{algorithm}[t]
    \caption{MIRA}
    \label{MIRA}
    \begin{algorithmic}[1]
        \STATE {\bf Parameters:} $\eta$, $(T, f, R, \mathcal{N}, \bm{W}^{(0)})$
        \STATE {\bf Initialize:} $\bm{B}, \bm{A}$


        
        \FOR{communication round $t = 1$ to $T$}

        \STATE Sample  a subset $S^{(t)}$ to initiate local training
        
            \FOR{each client $k \in S^{(t)}$}
            \IF{$t > 1$}
                \STATE Get latest model $\Delta \bm{W}_{k}^{(t)}$ from the server.
                \STATE $\Delta \bm{W}_{k, 0}^{(t)} = \Delta \bm{W}_{k}^{(t)}$
            \ENDIF
            
                \FOR{each local round $r= 1$ to $R$}

                \STATE $\Delta \bm{W}_{k, r+1}^{(t)} = $\textbf{InstructionTuning}($\Delta \bm{W}_{k, r}^{(t)}$)
                
                \ENDFOR
                \STATE Send $\Delta$$\bm{W}_{k, R}^{(t)}$ to the server.                
            \ENDFOR

            \textbf{On Server}\\
            \FOR{each client $k \in S^{(t)}$}
                \STATE  $\Delta \bm{W}_k^{(t+1)} =\Delta \bm{W}_{k, R}^{(t)}-\eta \lambda \sum_{\ell \in \mathcal{N}_k} a_{k \ell}\left(\Delta \bm{W}_{k, R}^{(t)}-\Delta \bm{W}_{\ell, R}^{(t)}\right)$
          
            \ENDFOR

        \ENDFOR
    \end{algorithmic}
\end{algorithm}

\section{Proposed Algorithm}\label{METHOD}
In this section, we propose MIRA, a novel algorithm that enables fine-tuning LLMs in an FMTL manner with efficiency in terms of \textit{computation} and \textit{communication} costs. Our approach leverages LoRA as a PEFT method, which solves, to some extent, the memory footprint issue by reducing the number of trainable parameters in the model. We assume that each client is equipped with a pre-trained model $\bm{W}^{0} \in \mathbb{R}^{d \times v}$. This weight matrix is then decomposed into two smaller matrices such that 
\begin{align}
    \bm{W}^{0} + \Delta \bm{W} = \bm{W}^{0} + \bm{B}\bm{A},
\end{align}
where $\Delta \bm{W}$ represents the accumulated gradient updates, $B \in \mathbb{R}^{d \times r}$ and $\bm{A} \in \mathbb{R}^{r \times v}$, where $r$ being the rank of both $\bm{A}$, and $\bm{B}$. Instead of optimizing the original weights $\bm{W}^{0}$, we only update $\bm{A}$ and $\bm{B}$. LoRA hypothesises that projecting LLMs into lower-dimensional spaces preserves performance, allowing us to optimize these smaller matrices efficiently. In the rest of the paper, we refer to $\bm{B}$ and $\bm{A}$ as $\Delta \bm{W}$.
\begin{figure*}[t]
    \centering
    \begin{subfigure}[t]{0.4\textwidth}
        \includegraphics[width=0.95\textwidth]{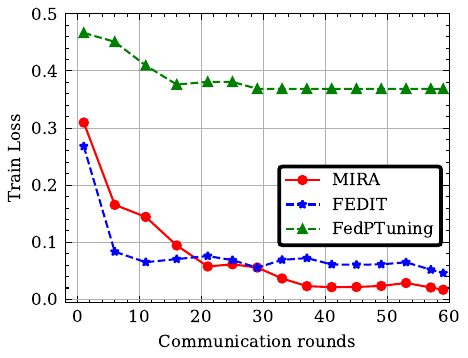}
        \caption{}
        \label{fig:figure1}
    \end{subfigure}
    \begin{subfigure}[t]{0.4\textwidth}
        \includegraphics[width=0.9\textwidth]{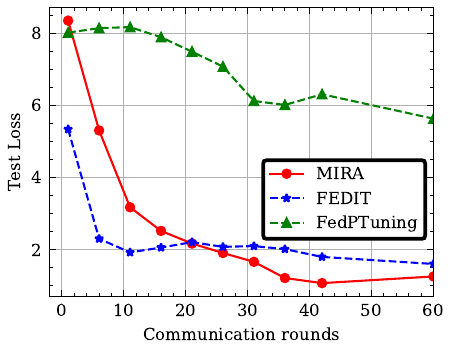}
        \caption{}
        \label{fig:figure2}
    \end{subfigure}
    \caption{Performance comparison of the proposed method and the baselines on Data-Juicer using the Natural Instruction dataset.}
    \label{fig:two_figures}
\end{figure*}
At communication round $t$, the server uniformly selects a subset of clients $S^{(t)}$. Each client, $k \in S^{(t)}$, performs $R$ local rounds of updates on the lower-dimensional matrix, $\Delta \bm{W}_k$, such that 
\begin{align}
\Delta \bm{W}_{k, R}^{(t)} = \textbf{InstructionTuning}(\Delta \bm{W}_{k, r}^{(t)}), r \in \{1, \dots,R\},
\end{align}
where \textbf{InstructionTuning}($\cdot$) refers to the process of adapting the weight matrix to the new dataset by computing a forward pass, and a backward pass, and then updating the model's weights. By leveraging LoRA, our forward pass becomes
\begin{align}
    \bm{h} = \bm{W}^{0} \bm{x} + \bm{B}\bm{A} \bm{x},
\end{align}
where $\bm{h}$ is the output of the forward pass, and $\bm{x}$ is a sample drawn from the dataset which is of the form of instructions $\bm{X}_k$ and the expected response from the model $\bm{y}_k$. After the forward pass, a backward pass is calculated with BP, computing the gradient w.r.t the LoRA's weights. Finally, we update the model weights using an optimizer step, yielding a locally updated model. This process continues for $R$ local rounds. The final locally updated model $\Delta \bm{W}_{k, R}^{(t)}$ is sent to the server. The server then performs a \textit{regularization update} to align clients with similar tasks, encouraging their models to be closer to each other. The server updates each client's matrix, $\forall k \in S^{(t)}$, as follows
\begin{multline}\label{reg.update}
     \Delta \bm{W}_k^{(t+1)} =\Delta \bm{W}_{k, R}^{(t)}-\eta \lambda \sum_{\ell \in \mathcal{N}_k} a_{k \ell}\left(\Delta \bm{W}_{k, R}^{(t)}-\Delta \bm{W}_{\ell, R}^{(t)}\right).
\end{multline} 
It is worth noting that the performance of the traditional aggregation scheme that averages the model's parameters might not be suitable due to the heterogeneity of the tasks. The update for non-sampled clients is given by
\begin{align}
    \Delta \bm{W}_k^{(t+1)} =\Delta \bm{W}_{k}^{(t)}, \forall k \notin S^{(t)}.
\end{align}
The process continues for $T$ communication rounds until convergence as depicted in Algorithm \ref{MIRA}.

\section{Experiments}\label{RES}
In this section, we discuss the details of our experiments and compare the performance of our approach, MIRA, to baseline schemes of existing federated fine-tuning methods for LLMs. Afterwards, we compare the computational and communication overhead attributed to MIRA, in comparison to the baseline schemes. Finally, we analyze the effect of FMTL on the overall per-task performance.
\subsection{Experimental Setup}
We conduct our experiments using two models, Data-Juicer~\cite{chen2023data}, a 1.5 billion parameters Llama-based model and GPT-2-large~\cite{radford2019language}, which has around $124$ million parameters. We utilize two datasets, namely NI~\cite{wang2022super} and Dolly-15k~\cite{DatabricksBlog2023DollyV2}. The number of communication rounds for all experiments is set to $60$.
We follow the same pre-processing steps described by~\cite{qin2023federated} and only sample around 20\% of the training set and $2\%$ of the test set. 
For Dolly-15k and Natural Instruction, we consider a scenario with $80$ clients, where every client has a unique local task $t$. 
At each communication round, we sample $10\%$ of all clients to participate in the round.
The metrics we utilize to assess the performance are the average train and test losses and the Rouge-L\cite{lin2004rouge}, a metric that is widely adopted to measure the quality of LLM's output. Rouge-L measures the longest matching sequence of words, i.e., longest common subsequence (LCS), between a candidate text and a reference, assessing the quality of text generation.
The batch size is set to $8$ to guard against any memory limitations and the rank of the LoRA matrices is set to $16$. we set the adjacency matrix $A$'s values randomly.
All Algorithms are tested and run on an A100 GPU equipped with a total of $40$ GB VRAM. We compare our method to two other baseline schemes that are tailored specifically for federated fine-tuning of LLMs, namely FedIT~\cite{10447454} and FedPTuning~\cite{kuang2023federatedscopellm}. The best hyperparameters for all three methods are selected for all methods.
\subsection{Performance Comparison}
In Fig. \ref{fig:two_figures}, we compare the performance of our proposed scheme, MIRA, against the baselines. For Natural Instruction with Data-Juicer, it is observed that MIRA outperforms the two baselines. In fact, it takes around $20$ communication rounds for MIRA to outperform FedIT, the closest baseline. The Rouge-L scores of the three methods are reported in Table~\ref{tab2} for the Dolly-15k dataset. We can see that MIRA significantly outperforms FedPTuning leading to an improvement in the Rouge-L scores by $6.1\%$ and $14.3\%$, for the Data-Juicer and GPT2-Large models, respectively. On the other hand, MIRA remains highly competitive with FedIT achieving comparable Rouge-L scores.  

%
\subsection{Memory and Communication Costs}
We study the computation and communication costs of our approach and the baselines. 
For communication cost, we quantify the number of communicated bits from the clients to the server and vice versa. We denote the upload and download link communication costs as $U$ and $D$, respectively. 
Then, the total communication cost is $U + D$. 
The memory cost is estimated as a function of the original model size $C_w$, the size of the trainable parameters $\hat{C}_{\Delta w}$ and any other overhead that results from saving the optimizer's states $O$. The overall memory cost is then $C_w + \hat{C}_{\Delta w} + O$. We report the cost of one round per client. As we can see in Table \ref{tab1}, MIRA exhibits the same communication and memory costs as FedIT, given that the same rank is used by both of them to obtain the PEFT model. On the other hand, FedPTuning is lighter compared to them consuming less communication and memory costs.
\begin{table}
\caption{Rouge-L values of the different algorithms on Dolly-15k dataset.}
\begin{center}
\begin{tabular}{@{}lcc@{}}
\toprule
\textbf{Model} &  \textbf{Algorithm} & \textbf{Rouge-L}(\%)\\
\midrule

 & MIRA& \textbf{20.8}\\
Data-Juicer& FedIT&19.1 \\
 & FedPTuning~& 14.7\\

\hline
& MIRA& 26.6\\
GPT2-large & FedIT& \textbf{28.4}\\
 & FedPTuning& 12.3\\
 
\bottomrule

\end{tabular}
\label{tab2}
\end{center}
\end{table} 
\begin{table}
\caption{Communication and memory costs of the different algorithms.}
\begin{center}
\begin{tabular}{@{}cccc@{}}
\toprule
\textbf{Model} &  \textbf{Algorithm} & \textbf{Communication cost}& \textbf{Memory Cost}\\
\midrule

 & MIRA& 12 MB &2.60 GB\\
Data-Juicer & FedIT& 12 MB &2.60 GB\\
 & FedPTuning&2.22 MB &2.57 GB\\

\hline
& MIRA&7.74 MB &1.51 GB\\
GPT2-large& FedIT&7.74 MB &1.51 GB \\
 & FedPTuning& 0.71 MB&1.48 GB\\
 
\bottomrule

\end{tabular}
\label{tab1}
\end{center}
\end{table}

\subsection{Per-Task Performance Analysis}
Finally, we analyze the effect of FMTL on the overall per-task performance to understand how well MIRA adapts to individual tasks compared to the baselines. We randomly select four clients/tasks and report the local test loss. As observed in Table~\ref{tab3}, MIRA achieves a lower average loss per client/task in three out of the four clients, which indicates the effectiveness of FMTL and its adaptability to task-specific requirements in contrast to the model averaging scheme of FedIT and FedPTuning.

\section{Conclusion}
In this work, we introduced MIRA, a federated finetuning paradigm for LLMs. MIRA utilizes Multi-Task Learning to enhance the models and make them personalized while leveraging other clients' models. By utilizing LoRA for parameter-efficient finetuning, MIRA maintains model adaptability while minimizing communication overhead. The superior performance of MIRA can be attributed to its effective alignment of client models based on task similarity, facilitated by the regularization term in the objective function. 

\begin{table}
\caption{Local test loss per client for some selected subset of clients on Data-Juicer using the Natural Instruction dataset.}
\begin{center}
\begin{tabular}{@{}lcc@{}}
\toprule
\textbf{Client Task} &  \textbf{Algorithm} & \textbf{Test Loss}\\
\midrule

 & MIRA& 2.7\\
Question Answering&FedIT&\textbf{1.8}\\
&FedPTuning&9.55 \\
\hline
 & MIRA&\textbf{0.23} \\
Program Execution&FedIT&0.37\\
&FedPTuning&0.87\\
\hline
 & MIRA& \textbf{1.87}\\
Speaker Identification&FedIT&7.01\\
&FedPTuning&11.98 \\
\hline
 & MIRA&\textbf{1.99 }\\
Explanation&FedIT&2.55\\
&FedPTuning&3.08\\

\bottomrule

\end{tabular}
\label{tab3}
\end{center}
\end{table}



\bibliographystyle{IEEEtran}
\bibliography{references} 

\end{document}